\documentclass[journal,twocolumn]{IEEEtran}
\usepackage[colorlinks,
            linkcolor=blue,
            anchorcolor=blue,
            citecolor=green]{hyperref}
\usepackage{amsmath}
\usepackage{subfigure}
\usepackage{caption}
\usepackage{algorithm}
\usepackage{algpseudocode}
\usepackage{setspace}
\usepackage{color,hyperref}

\usepackage{graphicx}
\usepackage{amsmath,amssymb} 
\usepackage{epsfig}
\usepackage{amsmath}
\usepackage{amssymb}
\usepackage{colortbl}
\usepackage{soul}
\usepackage{multirow}
\usepackage{pifont}
\usepackage{url}
\usepackage{soul}
\usepackage[table]{xcolor}

\ifCLASSINFOpdf
\else
\fi

\definecolor{darkblue}{rgb}{0.0,0.0,1.0}
\hypersetup{colorlinks,breaklinks,
	linkcolor=darkblue,urlcolor=darkblue,
	anchorcolor=darkblue,citecolor=darkblue}

\definecolor{mygray}{gray}{.9}
\definecolor{Lavender}{HTML}{E6E6FA}
\definecolor{LightCyan}{HTML}{E1FFFF}
\newcommand{\xmark}{\ding{55}}
\newcommand{\rmark}{\ding{52}}

\newcolumntype{I}{!{\vrule width 1.2pt}}
\newlength\savedwidth
\newcommand\whline{\noalign{\global\savedwidth\arrayrulewidth
		\global\arrayrulewidth 1.25pt}%
	\hline
	\noalign{\global\arrayrulewidth\savedwidth}}

\begin{document}

\title{Neuron Linear Transformation: Modeling the Domain Shift for Crowd Counting}
\author{Qi~Wang,\IEEEmembership{~Senior Member,~IEEE}, Tao~Han,\IEEEmembership{~Student Member,~IEEE}, Junyu~Gao,\IEEEmembership{~Member,~IEEE}, and Yuan~Yuan,\IEEEmembership{~Senior Member,~IEEE}
	\thanks{	
	The authors are with the School of Computer Science and the Center for
	OPTical IMagery Analysis and Learning (OPTIMAL), Northwestern Polytechnical
	University, Xi'an 710072, China (e-mail: crabwq@gmail.com; hantao10200@mail.nwpu.edu.cn, gjy3035@gmail.com;  y.yuan@nwpu.edu.cn). Yuan Yuan is the corresponding author.

	\copyright 20XX IEEE. Personal use of this material is permitted. Permission from IEEE must be obtained for all other uses, in any current or future media, including reprinting/republishing this material for advertising or promotional	purposes, creating new collective works, for resale or redistribution to servers or lists, or reuse of any copyrighted component of this work in other works.}
}
\markboth{{IEEE} Transactions on Neural Networks and Learning Systems}%
{Shell \MakeLowercase{\textit{et al.}}: Bare Demo of IEEEtran.cls for Journals}
\maketitle


\begin{abstract}
Cross-domain crowd counting (CDCC) is a hot topic due to its importance in public safety. The purpose of CDCC is to alleviate the domain shift between the source and target domain. Recently, typical methods attempt to extract domain-invariant features via image translation and adversarial learning. When it comes to specific tasks, we find that the domain shifts are reflected on model parameters' differences. To describe the domain gap directly at the parameter-level, we propose a Neuron Linear Transformation (NLT) method, exploiting domain factor and bias weights to learn the domain shift. Specifically, for a specific neuron of a source model, NLT exploits few labeled target data to learn domain shift parameters.  Finally, the target neuron is generated via a linear transformation. Extensive experiments and analysis on six real-world datasets validate that NLT achieves top performance compared with other domain adaptation methods. An ablation study also shows that the NLT is robust and more effective than supervised and fine-tune training. Code is available at:
\url{https://github.com/taohan10200/NLT}.
\end{abstract}
\section{Introduction}
Currently, accelerating the crowd understanding is playing an increasingly important role in building an intelligent society. As a huge research field, it involves many hotspots. In some scenes with sparse crowd distribution, crowd understanding mainly includes crowd detection \cite{idrees2015detecting}, groups analysis \cite{C30}, crowd segmentation \cite{heim2017clickstream}, and crowd tracking \cite{gao2017beyond}. In some scenes with dense crowds, such as an image containing thousands of people, crowd understanding mainly focuses on counting and density estimation \cite{huang2017body,sindagi2019ha,ling2019indoor,tian2019padnet,jiang2019learning}. In this paper, we strive to work on the existing crowd counting problem.

\begin{figure}[t]
	\centering
	\includegraphics[width=.46\textwidth]{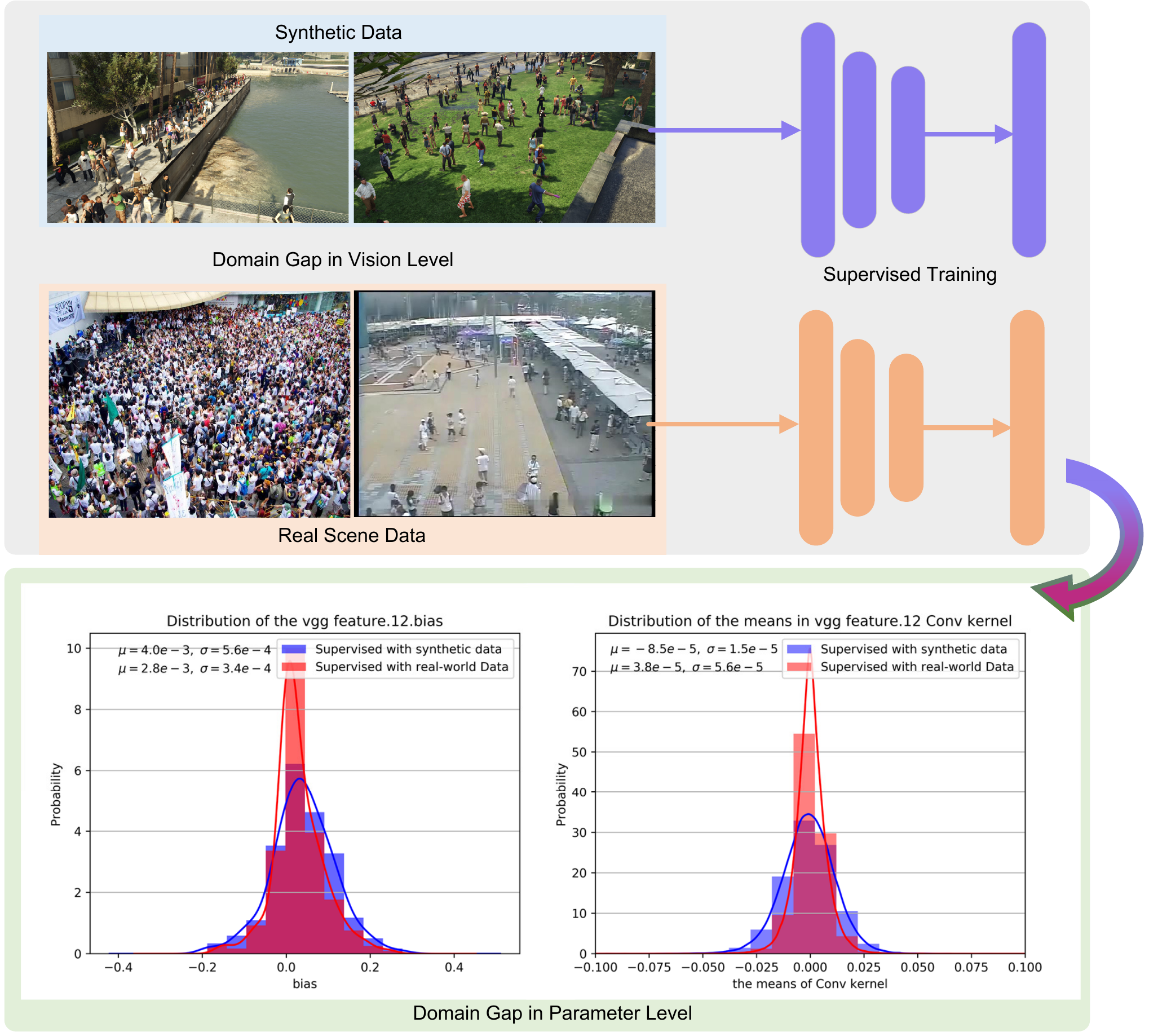}
	\caption{The domain shift in different views. 1) visual domain shift, such as brightness, background, character feature, etc. 2) when it comes to specific tasks, the domain shift is reflected in the model's parameter distribution.}\label{fig:domain}
\end{figure}
Crowd counting, a system that generates a pixel-level density estimation map and sums all of the pixels to predict how many people are in an image, has become a prevalent task due to its widespread piratical application: public management, traffic flow prediction, scene understanding \cite{zhao2019property}, video analysis \cite{zhao2019cam}, \emph{etc}. Specifically, it can be used for public safety in many situations, such as political rallies and sports events \cite{saleh2015recent}. Besides, density estimation can also help crowd localization in some sparse scenes \cite{idrees2018composition}. In the traditional supervised learning, many excellent algorithms \cite{sindagi2019ha,ma2019bayesian,zhao2019leveraging,wan2019residual,wan2019adaptive} constantly refresh the counting metrics from different angles for the existing datasets.

However, traditional supervised learning requires a lot of labeled data to drive it, and unfortunately, pixel-level annotating is often costly. According to statistics \cite{idrees2018composition}, the entire procedure involves 2, 000 human-hours spent in completing the QNRF dataset \cite{idrees2018composition}. For the recently established NWPU dataset \cite{wang2020nwpu}, the time cost is even as high as 3,000 human-hours. Even if researchers devote a lot of time and money to construct the datasets, the existing datasets are still limited in scale.

Because of the small-scale in some existing datasets, the above models may suffer from overfitting at different extents. It causes a significant performance reduction when applying them in real life. Thus, CDCC attracts the researcher's attention, which focuses on improving the performance in the target domain by using the data from the source domain.
Wang \emph{et al.} \cite{wang2019learning} propose a crowd counting via domain adaptation method, SE CycleGAN, which translates synthetic data to photo-realistic scenes, and then apply the trained model in the wild. Gao \emph{et al.} \cite{gao2019domain} present a high-quality image translation method feature disentanglement. \cite{han2020focus,gao2019feature} adopt the adversarial learning to extract the domain-invariant features in the source and target domain. In a word, general Unsupervised Domain Adaption (UDA) methods concentrate on image style and feature similarity. The upper box in Fig. \ref{fig:domain} demonstrates the appearance differences.

Nevertheless, the domain shift in image and feature level is not sensitive to the counting task: this strategy does not directly affect the counting performance, and it is not optimal. For example, SE CyCleGAN \cite{wang2019learning}, and DACC \cite{gao2019domain} focus on maintaining the local consistency to improve the translation quality in congested regions. When applying the model to the sparse scenes (Mall \cite{chen2012feature}, UCSD \cite{chan2008privacy}), the loss may be redundant. In other words, there are task gaps in the existing UDA-style methods. Besides, since the target label is unseen for UDA models, they do not work well, such as coarse prediction in the congested region and the estimation errors in the background.

Given a specific task, we find that the parameters' difference between two models can reflect the domain shift. Notably, we use synthetic data and real-scene data to train the model, respectively. Then, we calculate the mean value of each kernel in a specific layer. The bottom box in Fig. \ref{fig:domain} reports the distribution histogram. It can intuitively see that the parameters both show Gaussian distribution, and the differences are their mean and variance. Thus, we assume that the difference in parameter distribution can be exploited to measure the domain shift in two datasets.

\begin{figure*}[htbp]
\centering
\includegraphics[width=0.9\textwidth]{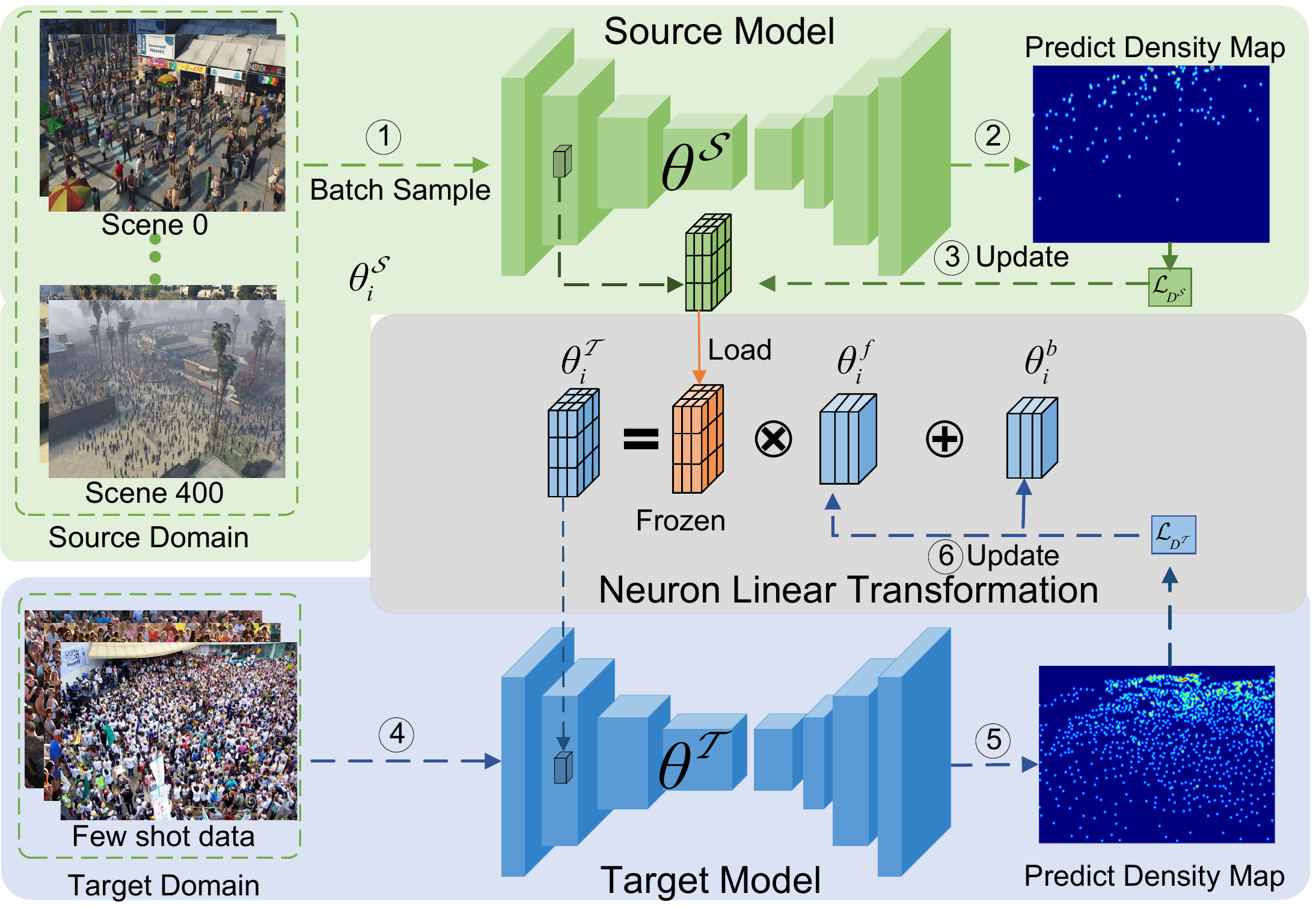}
\caption{The flowchart of our proposed NLT, which consists of three components: 1) Source model is trained with the synthetic data; 2) The learnable parameters $\theta^{f}$ and $\theta^{b}$ are used to model the domain shift, which are defined according the source model. Namely,  for a source neuron $\theta_{i}^{\mathcal{S}} \in \theta^{\mathcal{S}}$ ($i$ is the index of neurons), there are a $\theta_{i}^{f}$ and a $\theta_{i}^{b}$ that are used to generate a target neuron $\theta_{i}^{\mathcal{T}}$ by a linear operation. 3) After loading the transferred parameters $\theta^{\mathcal{T}}$ to the target model, the few-shot data are feed into the target model to update the domain shift parameters.}
\label{fig:framework}
\end{figure*}

According to the above observation, the domain shift on the parameter level can be simulated by a linear transformation. Thus, this paper proposes an NLT method to tackle cross-domain crowd counting. To be specific, firstly, train a source model with traditional supervised learning on synthetic data. Then, exploit a few labeled target data to learn two parameters (factor and bias) for each source neuron. Finally,  generate target model neurons from source neurons with a linear transformation. The entire process is shown in Fig. \ref{fig:framework}.

In summary, the main contributions of this paper are:
\begin{itemize}
  \item Propose a novel Neuron Linear Transformation method to model the domain shift. It is the first time that the domain shift can be measured at the parameter level.
  \item Design a newly few-shot learning framework to optimize the domain shift parameters, while few-shot learning in other CDCC methods is exploited to fine-tune the partial layers.
  \item Achieve more practical results on adapting synthetic dataset to six real-word crowd counting datasets. Further experiments show that NLT can also promote supervised learning performance.
\end{itemize}

\section{Related Work}

In this section, we briefly review the relevant works from the three tasks: supervised crowd counting, cross-domain crowd counting, and few-shot learning.

\textbf{Supervised Crowd Counting.}\quad
Most of the supervised crowd counting algorithms focus on addressing scale variability in recent years. From the multi-column scale-aware architecture, zhang \emph{et al.} \cite{zhang2016single} propose a three-columns network with different kernels for scale perception. L{\'o}pez-Sastre \emph{et al.} \cite{onoro2016towards} introduce a HydraCN with three-columns, where each column is fed by a patch from the same image with a different scale. Wu \emph{et al.} \cite{wu2019adaptive} develop a powerful multi-column scale-aware CNN with an adaptation module to fuse the sparse and congested column. To generate a high-quality density map, AFP \cite{kang2018crowd} fuses the attention map and intermediary density map in each column. ic-CNN \cite{ranjan2018iterative} delivers the feature and predicted density map from the low-resolution CNN to the high-resolution CNN. Hossain \emph{et al.} \cite{hossain2019crowd} propose a multi-column scale-aware attention network, where each column is weighted with the output of a global scale attention network and local scale attention network. Besides, the single-column scale-aware CNN \cite{cao2018scale,zhang2018crowd} also make great progresses in recent research. CSRNet\cite{li2018csrnet}, CAN \cite{liu2019context}, and FPNCC \cite{cenggoro2019feature} employ multiple paths only in partly layers, which is a combination of multi-column and single-column scale-aware CNN.

From other perspectives, HA-CNN \cite{8767009} employs a spatial attention module (SAM) and a set of global attention modules (GAM) to enhance the features extracting ability selectively. To further refine the density map, CRNet \cite{9096602} stacks several fully convolutional networks together recursively with the previous output as the next input. Every stage utilizes previous density output to refine the predicted density maps gradually. To count people in various density crowds, PaDNet \cite{8897143} develops three components: Density-Aware Network (DAN) module extracts pan-density information, Feature Enhancement Layer (FEL) effectively captures the global and local contextual features, and  Feature Fusion Network (FFN) embeds spatial context and fuses these density-specific features. CLPNet \cite{8798674} exploits a cross-level parallel network to conduct multi-scale fusion from five different aggregation modules. It extracts multiple low-level features from VGG-16 and then fuses them with specific scale aggregation modules in the high-level stage. To tackle the crowd distribution and background interference problems, Mo \emph{et al.} \cite{9161353} utilize local information of distance between human heads and the global information of the people distribution in the whole image to achieve head size estimation. The predicted head masks are used to reduce the impact of different crowd scale and background noise.

\textbf{Cross-domain Crowd Counting.}\quad
In addition to the exploration mentioned above, CCDC, a new research hotspot, is beginning to interest researchers. This task aims to transfer what the model learns from one dataset to another unseen dataset. Literature \cite{wang2019learning} is the earliest research in this filed, which establishes a large-scale synthetic dataset to pre-train a model that improves the robust over real-world datasets by fine-tuning. Except fine-tuning, it also trains a counter without using any real-world labeled data. It narrows the domain gap by using the Cycle GAN \cite{zhu2017unpaired} and SE Cycle GAN \cite{wang2019learning,wang2020pixel} to generate a realistic image. Recently, several efforts have been made to follow it, DACC \cite{gao2019domain}, a method for domain adaptation based on image translation and Gaussian-prior reconstruction, achieves new state-of-the-art results on several mainstream datasets. At the same time, some works \cite{gao2019feature,han2020focus} extract domain invariant features based on adversarial learning. Experimental results show that those methods can narrow the domain shift to some extent.

For the supervised crowd counting, the advantage is that it can get well-performed models by conventional training. However, the supervised methods require much manual annotation samples, which are laborious to get them. Besides, the model trained in a specific scenario does not work well when applying to other scenarios because of the domain gap. CDCC can alleviate the above shortcomings, but the current CDCC methods that adapt to real-world scenes from synthetic data cannot achieve similar performance with supervised learning. Overall, the intersection of synthetic data and real-world data proves to be particularly fertile ground for groundbreaking new ideas, and this field will become more significant over time.

\textbf{Few-shot Learning.}\quad
Since it involves a small number of target domain samples in our CDCC method, we hereby introduce some studies related to few-shot learning. The few-shot learning is based on prior experience with very similar tasks where we have access to large-scale training sets and then train a deep learning model using only a few training examples. Early few-shot learning methods \cite{bart2005cross,fei2006knowledge,fink2005object} are based on hand-crafted features. Vinyals \emph{et al.} \cite{vinyals2016matching} use a memory component in a neural net to learn common representation from very little data. Snell \emph{et al.} \cite{snell2017prototypical} propose Prototypical Networks, which map examples to a dimensional vector space. Ravi and Larochelle \cite{ravi2016optimization} use an LSTM-based meta-learner to learn an update rule for training a neural network learner. Model-Agnostic Meta-Learning (MAML) \cite{finn2017model} learns a model parameter initialization that generalizes better to similar tasks. Similar to MAML, Mishra \emph{et al.} \cite{nichol2018first} executes stochastic gradient descent for $K$ iterations on a given task, and then gradually moves the initialization weights in the direction of the weights obtained after the $K$ iterations. Santoro \emph{et al.} \cite{santoro2016meta} propose Memory-Augmented Neural Networks (MANNs) to memorize information about previous tasks and leverage that to learn a learner for new tasks. SNAIL \cite{mishra2017simple} is a generic meta-learner architecture to learn a common feature vector for the training images to aggregate information from past experiences. Most of the above few-shot learning methods are based on classification tasks.
Besides, few-shot learning has been applied to many computer vision tasks. Siamese-based trackers can be viewed as an application of one-shot learning \cite{Fully2016,NIPS2016,Dynamical2019,Dong_2018_ECCV,Quadruplet,Dong_2018_CVPR}. For example, \cite{NIPS2016} get a learner net by off-line training, and then generate a parameter of pupil net online with one sample. \cite{Quadruplet} propose a new quadruplet deep network that contains four branches of the same network with shared parameters. It achieves a more powerful representation by examining the potential connections among the training instances.
For crowd counting tasks, \cite{hossainone} proposes a one-shot learning approach for learning how to adapt to a target scene using one labeled example. \cite{reddy2020few} applies the MAML \cite{finn2017model} to learn scene adaptive crowd counting with few-shot learning.

\section{Approach}
In this section, we first define the problem that we want to solve. Then, the NLT, a linear operation at the neuron level, is designed to model the domain shift. Finally, we introduce how to transfer the source model to the target model with NLT operation. Fig. \ref{fig:framework} illustrates the entire framework.
\subsection{Problem Setup}
In this paper, we strive to tackle the existing problems for domain adaptive crowd counting from the parameter-level with a transformation. The setting assumes access to a source domain (synthetic data) with $N_{\mathcal{S}}$ labeled crowd images $D^{\mathcal{S}}=\{I_{i}^{\mathcal{S}}, Y_{i}^{\mathcal{S}}\}_{i}^{N_{\mathcal{S}}}$. Besides, a target domain (real scene data) provides $N_{\mathcal{S}}$ few-shot images with the labeled density maps $D^{\mathcal{T}}=\{I_{i}^{\mathcal{T}}, Y_{i}^{\mathcal{T}}\}_{i}^{N_{\mathcal{T}}}$. The purpose is to train a source domain model $\mathcal{S}$  with the parameters $\theta^{\mathcal{S}}$ exploiting the $D^{\mathcal{S}}$, and learn a representable domain shift according to $D^{\mathcal{T}}$ with few-shot learning, which are parameterized by the domain factors $\theta^{f}$ and domain biases $\theta^{b}$. Finally,  generating a well performed target model $\mathcal{T}$ with the parameters $\theta^{\mathcal{T}}$ by combining the source model with the domain shift parameters.

\subsection{Neuron Linear Transformation}
\label{neuron:define}
\begin{figure}[htbp]
\centering
\includegraphics[width=0.48\textwidth]{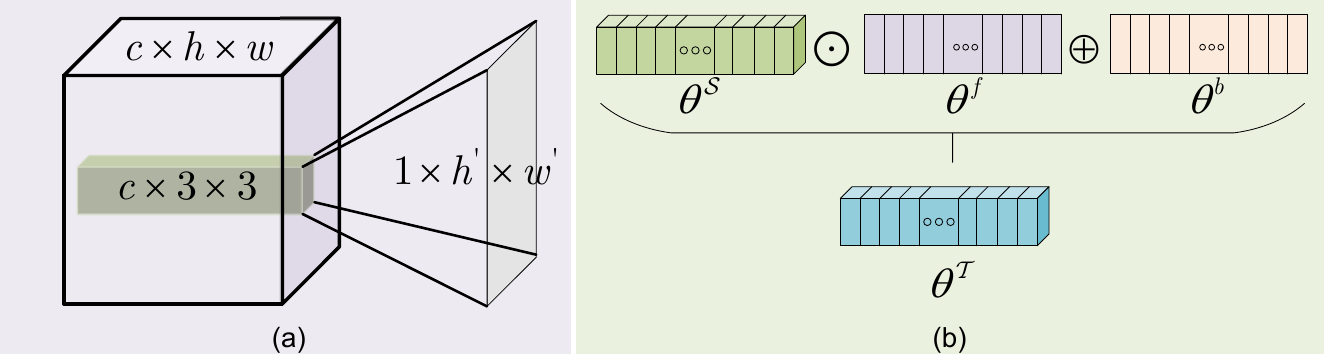}
\caption{(a) is the schematic diagram of neuron's definition. (b) shows how to transfer source neuron $\theta^{\mathcal{S}}$ to target neuron $\theta^{\mathcal{T}}$ with NLT.}
\label{fig:neuron}
\end{figure}

Inspired by the scale and shift operation \cite{sun2019meta}, we propose a Neuron Linear Transformation method to describe the domain gap, which makes the domain gap visible. To model the domain shift, we assume that  \emph{the source model and the target model belong to the same linear space $V^{n}$. Each neuron in the target model can be transferred from the corresponding neuron in the source model by a linear transformation.} As shown in Fig. \ref{fig:neuron} (a), a neuron is defined as a convolutional group used to generate a channel ($1\times h_{'} \times w_{'}$) in the next layer from the last layer ($c\times h \times w$). The size of neurons is determined by two factors: the number of channels in the last layer of the CNN  and the size of the convolution kernel used to generate the next layer. Fig. \ref{fig:neuron} (b) shows how the target model's parameters are transferred from the source model by a linear operation. For the source model's neuron $\theta^{\mathcal{S}}\in \mathbb{R}^{c\times h\times w}$, we define the corresponding domain factor $\theta^{f} \in \mathbb{R}^{c\times 1 \times 1}$ and domain bias $\theta^{b} \in \mathbb{R}^{c\times 1 \times 1}$. Then the neuron-level linear transformation can be formulated as follow.
\begin{equation}
\small
\begin{array}{l}
\begin{aligned}
\theta^{\mathcal{T}}=&\left[f^{1} \times\left[
\begin{array}{ccc}
{a_{11}^{1}} & {\cdots} & {a_{1 w}^{1}} \\
{\vdots} & {\ddots} & {\vdots} \\
{a_{1 h}^{1}} & {\cdots} & {a_{h w}^{1}}
\end{array}
\right]+\left[b^{1}\right], \cdots, \right.\\
& \left. f^{c} \times\left[
 \begin{array}{ccc}
{a_{11}^{c}} & {\cdots} & {a_{1 w}^{c}} \\
{\vdots} & {\ddots} & {\vdots} \\
{a_{1 h}^{c}} & {\cdots} & {a_{h w}^{c}}
\end{array}
\right]+b^{c}\right],
\end{aligned}\label{Equ:NLT}
\end{array}
\end{equation}
where $f^{i} \in \theta^{f}, b^{i} \in \theta^{b}$ and $a_{hw}^{i} \in \theta^{\mathcal{S}} \quad (i=1,2,\cdots,c)$. The domain adaptation method has two advantages: 1) The target model inherits the good feature extraction ability from the source model and preserves the generalization. 2) Compared with fine-tuning operation, fewer parameters need to be optimized in the target model with NLT. So it reduces the probability of overfitting for few-shot learning in the target domain.
\subsection{Modeling the Domain Shift}
Firstly, we introduce a crowd counter to test the NLT. Following the previous work \cite{wang2019learning,han2020focus,gao2019domain,gao2019feature}. we take the VGG architecture as the feature extractor. As shown in Fig. \ref{fig:framework}, the first ten layers of VGG-16 \cite{vgg16} are adopted as the backbone in the encoder stage. That is, the width and height of the output feature are 1/8 of the input image. In the decoder stage, a 3x3 convolutional layer is used to reduce the feature channels to a half. Then an up-sampling layer is followed by a 3x3 convolutional layer to reduce channels. After three repetitions, a 1x1 convolutional layer outputs the prediction density map. The source domain model's training is similar to that of the traditional supervised crowd counting network, except that the training data adopts the synthetic dataset. The $\theta^{\mathcal{S}}$ are optimized by gradient descent as follow,
\begin{equation}
\tilde{\theta}^{\mathcal{S}} =\theta^{\mathcal{S}}-\alpha \nabla \mathcal{L}_{D^{\mathcal{S}}} \left(\theta^{\mathcal{S}}\right),
\label{Equ:source}
\end{equation}
where $\mathcal{L}_{D^{\mathcal{S}}} \left(\theta^{\mathcal{S}}\right)=\frac{1}{2 n} \sum_{i}\left\|\mathcal{S}(I_{i}^{\mathcal{S}};\theta^{\mathcal{S}})-Y_{i}^{\mathcal{S}}\right\|_{2}^{2}$ is a standard MSE loss. $n$ is the batch size of source model. $\mathcal{S}(I_{i}^{\mathcal{S}};\theta^{\mathcal{S}})$ is the source model prediction of the $i^{th}$ training data. $\alpha$ denotes the learning rate.

Secondly, we introduce how to embed NLT into our target model training. As shown in Fig. \ref{fig:framework}, the target model keeps the same architecture as the source model, but the parameters in the target model are transferred from the source model. So, the number of learnable parameters is different. Taking the VGG-16 \cite{vgg16} backbone as an example, the channels of the first ten layers are $\{64,64,128,128,256,256,256,512,512,512\}$. According to the neuron's definition in \ref{neuron:define}, the total number of neurons is 2688, which is the sum of the channels' number. Assuming that the source model contains $k$ neurons, each neuron needs a factor and a bias in the target model. So the number of the learnable parameters $\theta^{f}$ and $\theta^{b}$ is $2\times k$. As the convolution kernel of VGG-16 is $3\times 3$, learnable parameters in the target model are $\sim2/9$ of the source model. The $\theta^{f}$ and $\theta^{b}$ are defined to model domain shift by neuron-level linear transformation. Specifically, we model the domain shift by transferring all neurons in the source model to the target model with the proposed NLT. According to Equ. (\ref{Equ:NLT}), the mapping can be expressed as follows,
\begin{equation}
\theta_{i}^{\mathcal{T}} =\theta_{i}^{\mathcal{S}} \odot \theta_{i}^{f} \oplus \theta_{i}^{b}, (i=1,2,\cdots,k),
\label{Equ:factor}
\end{equation}
where $\theta_{i}^{f}$ represents the domain shift factor, initialized by 1. $\theta_{i}^{b}$ represents the domain shift bias, which is initialized to 0.

Since we introduce the learnable parameters to describe the task gap, some target domain labeled images are needed to optimize the parameters. However, within the requirement of domain adaptation, we only use a few data to support the training. During training the source model, $\theta^{\mathcal{S}}$ are learned. However, they will be frozen when the target model is updated. After the calculation of Equ. \ref{Equ:factor},  $\theta^{\mathcal{T}}$ participate in the feed-forward of the target model. Therefore, only the gradients of $\theta^{f}$ and $\theta^{b}$ need to be calculated in the back-forward process. That is, $\theta^{f}$ and $\theta^{b}$ are learned in the target model. The loss for optimizing the parameters is defined as follows,
\begin{equation}
\small
\begin{array}{l}
\begin{aligned}
\mathcal{L}_{D^{\mathcal{T}}} \left(\theta^{\mathcal{T}}\right)= & \frac{1}{2 n} \sum_{i}\left\|\mathcal{T}(I_{i}^{\mathcal{T}};\theta^{\mathcal{T}})-Y_{i}^{\mathcal{T}}\right\|_{2}^{2} + \\
& \lambda (\sum_{i=1}^{k} (\theta_{i}^{f}-1)^{2} + (\theta_{i}^{b})^{2}),
\end{aligned}
\label{Equ:loss}
\end{array}
\end{equation}
where the former term is the density estimated loss of the few-shot data. It is the same as the loss of the source model. $ (I_{i}^{\mathcal{T}}, Y_{i}^{\mathcal{T}}) \in D^{\mathcal{T}}$  is the $i^{th}$ input image and density map. $\mathcal{T}(I_{i}^{\mathcal{T}};\theta^{\mathcal{T}})$ is the prediction density map. The latter term is the L2 regularization loss of parameters $\theta^{f}$ and $\theta^{b}$, with the purpose of preventing overfitting $D^{\mathcal{T}}$ in the target domain. $\lambda$ is the weighted parameter. Finally, the target model is optimized as follows,

\begin{figure*}[htbp]
\centering
\includegraphics[width=0.95\textwidth]{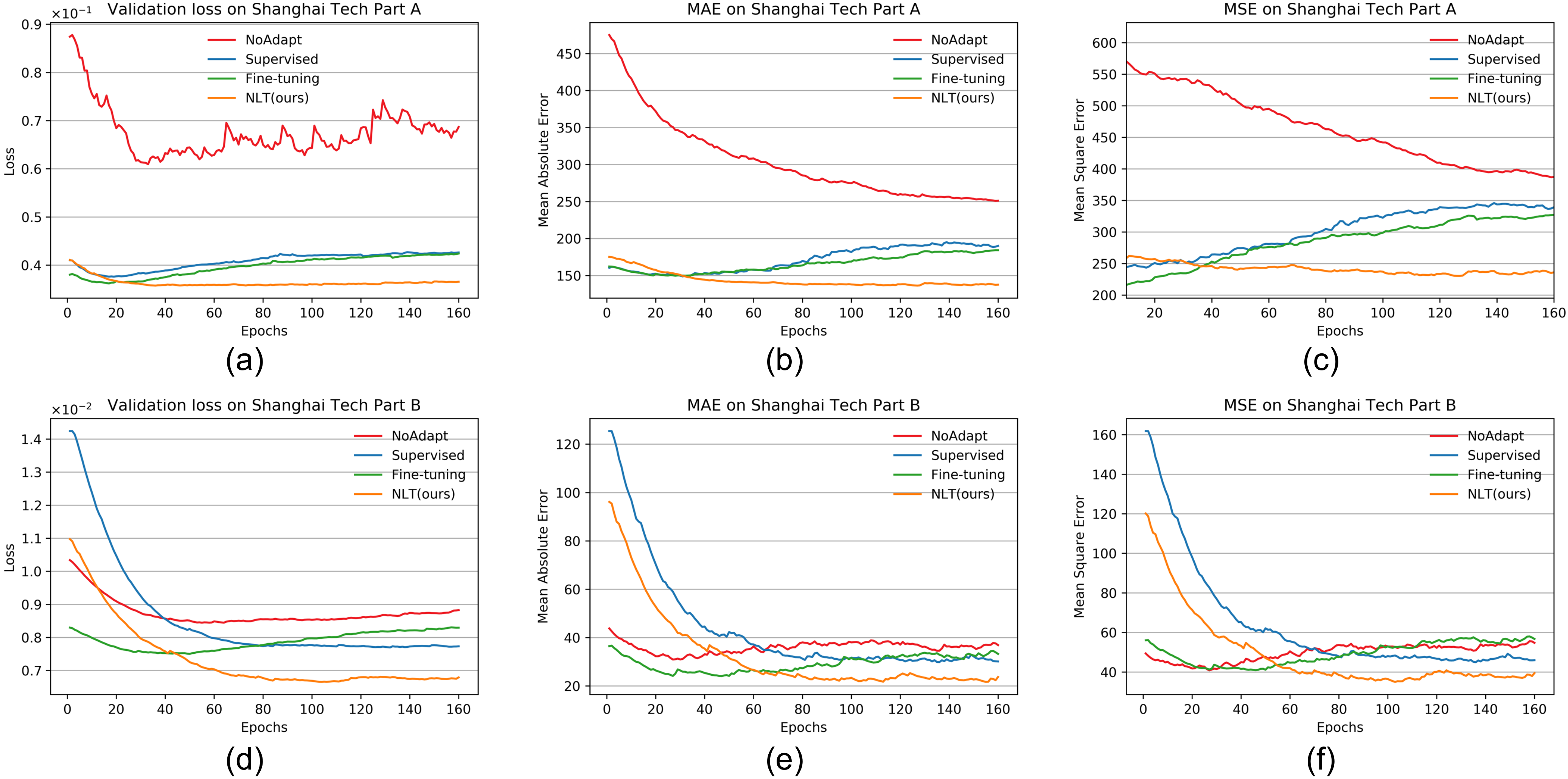}
\caption{The effects of our NLT and other training methods on learning process and performance. (a)(b)(c) and (d)(e)(f) show the validation loss and performance on Shanghai Tech Part A and B dataset, respectively.}
\label{fig:methods}
\end{figure*}

\begin{table*}[htbp]
	\centering
	\caption{The performance of different training methods on Shanghai Tech Part A and Shanghai Tech Part B.}
	\setlength{\tabcolsep}{8pt}{
		\begin{tabular}{c|ccIc|c|c|cIc|c|c|c}
			\whline
			\multirow{2}{*}{Method}	&\multirow{2}{*}{DA} &\multirow{2}{*}{FS} &\multicolumn{4}{cI}{Shanghai Tech Part A} &\multicolumn{4}{c}{Shanghai Tech Part B}\\
			\cline{4-11}
			&&& MAE &MSE &PSNR &SSIM &MAE & MSE &PSNR &SSIM \\
			\whline
			NoAdpt         &\xmark &\xmark &188.0 &279.6 &20.91 &0.670  &20.1 &29.2 &26.62 &0.895\\
			\hline
			Supervised     &\xmark &\rmark &107.2 &165.9 &21.53 &0.623   &16.0 &26.7 &26.8 &0.932 \\
			\hline	
			Fine-tuning a    &\rmark &\rmark &105.7 &167.6 &21.72 &0.702 &13.8 &22.3 &27.0 &0.931    \\
            \hline
            Fine-tuning b    &\rmark &\rmark &110.8 &167.4 &20.98 &0.722 &13.1 &20.8 &27.14 &0.924    \\
			\whline	
			NLT (ours)     &\rmark  &\rmark &\underline{93.8} &\underline{157.2} &\underline{21.89} &\underline{0.729}
                                            &\underline{11.8} &\underline{19.2}  &\underline{27.58} &\bfseries{0.937}  \\
            \hline
            IFS \cite{gao2019domain}+NLT (ours)     &\rmark  &\rmark &\bfseries90.1 &\bfseries151.6 &\bfseries22.01 &\bfseries0.741
                                            &\bfseries10.8 &\bfseries18.3  &\bfseries27.69 &\underline{0.932}  \\
			\whline
		\end{tabular}
		
	}
\label{Table:methods}
\end{table*}

\begin{equation}
\tilde{\theta}^{\mathcal{T}} =\theta^{\mathcal{T}}-\beta \nabla \mathcal{L}_{D^{\mathcal{T}}} \left(\theta^{\mathcal{T}}\right),
\end{equation}
where $\beta$ denotes the learning rate of target model.

\section{Implementation Details}
\textbf{Executive Stream.}\quad In the training phase, the workflow is shown in Fig. \ref{fig:framework} \textcircled{1} $\sim$ \textcircled{6}, once iteration updates parameters for two models. Firstly, as shown in \textcircled{1} $\sim$ \textcircled{3}, $\theta^{\mathcal{S}}$ are updated according to a batch sampling from the GCC dataset. Secondly, in \textcircled{4} $\sim$ \textcircled{6}, the domain shift parameters are updated with the few-shot data provided in the target domain. Finally, the parameters of the target model are obtained by NLT. In the testing phase, we use the best-performing model on the validation set to make an inference.

\textbf{Parameter Setting.}\quad In each iteration, we input $12$ synthetic images and $4$ target few-shot images. Adam algorithm \cite{kingma2014adam} is performed to optimize the networks. The learning rate $\alpha$ in the source model and $\beta$ in the target model are initialized as $10^{-5}$. The parameter $\lambda$ for target model loss function in Eq. \ref{Equ:loss} is fixed to $10^{-4}$. Our code is developed based on the $C^{3}$ Framework \cite{C3}.

\textbf{Scene Regularization.}\quad In some domain adaptation fields, such as semantic segmentation, the label distribution is highly consistent in two domains. Unlike that, current real-world crowd datasets are very different in density range, such as the number of the people in MALL \cite{chan2008privacy} dataset is ranging from $13$ to $53$, but the GCC \cite{wang2019learning} dataset is ranging from $0$ to $3,995$. For avoiding negative adaptation by the different density ranges, we adopt a scene regularization strategy proposed by \cite{wang2019learning} and  \cite{gao2019feature}. In other words, we add some filter conditions to select proper synthetic images from GCC as the source domain data for different real-world datasets.
\begin{figure*}
\centering
 \includegraphics[width=0.95\textwidth]{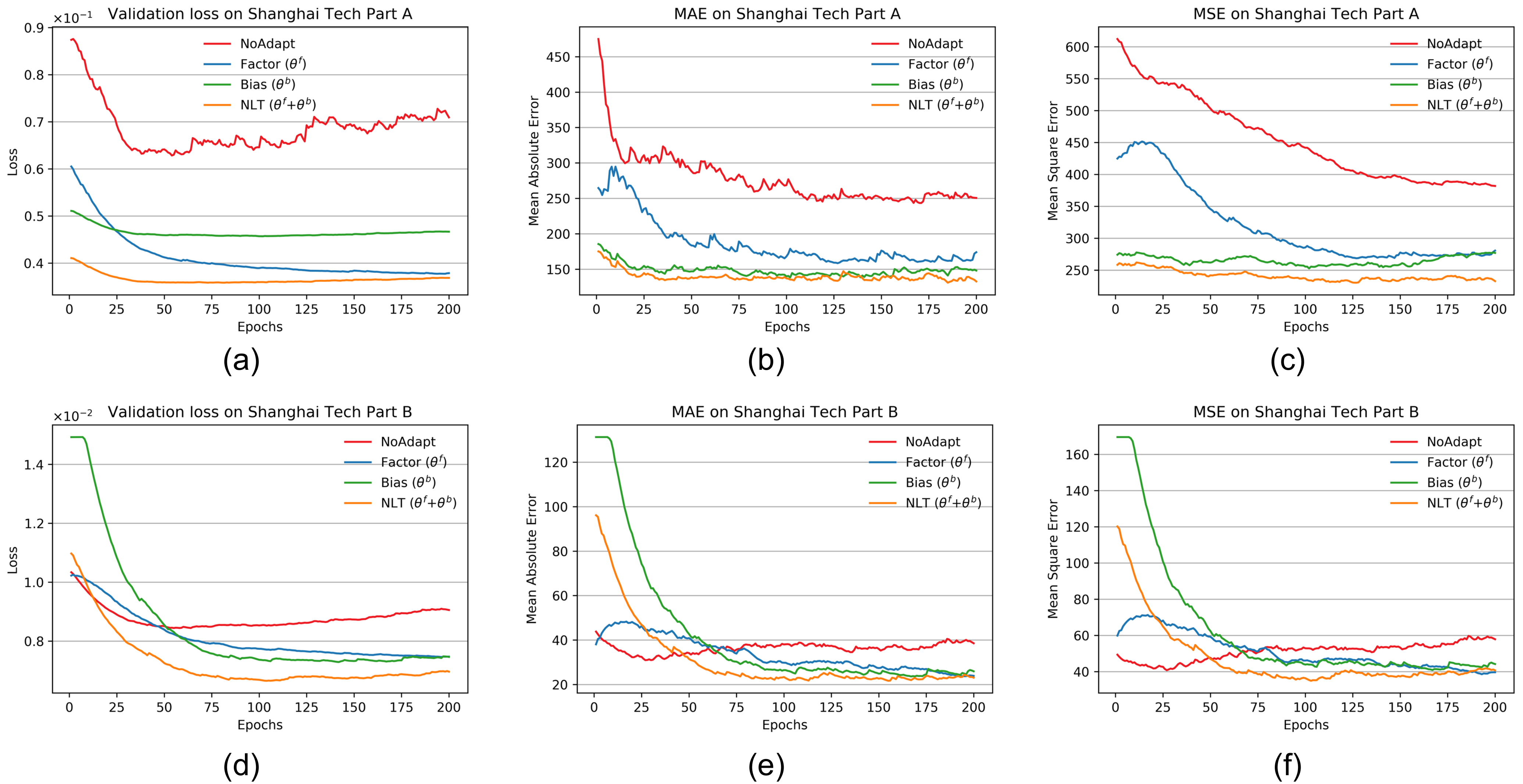}
\caption{The effects of the domain shift parameters $\theta^{f}$ and  $\theta^{b}$. (a)(b)(c) and (d)(e)(f) show the validation loss and performance on Shanghai Tech Part A and B dataset, respectively.}
\label{fig:parameters}
\end{figure*}

\section{Experiments}
In this section, we first introduce the evaluation metrics and the selected datasets, and then a comprehensive ablation study is performed to verify the effectiveness of our proposed method. Next, we analyze the shifting phenomenon in adopting synthetic dataset to different real-world datasets from statistics. Besides, we also discuss the effect of selected few-shot data on performance improvement. Finally, we present the testing results and visualization results of our method in six real-world datasets.

\subsection{Evaluation Criteria}
\textbf{Counting Error.}\quad According to the evaluation criteria widely used in crowd counting, the counting error is usually reflected in two metrics, namely Mean Absolute Error(MAE) and Mean Square Error(MSE). MAE measures the mean length of the predicted Error, while MSE measures the model's robustness to outliers. They are defined as follows:
\begin{equation}
\small
M A E=\frac{1}{N} \sum_{i=1}^{N}\left|y_{i}-\hat{y}_{i}\right|, M S E=\sqrt{\frac{1}{N} \sum_{i=1}^{N}\left|y_{i}-\hat{y}_{i}\right|^{2}},
\end{equation}
where $N$ is the number of images to be tested, and $y_{i}$ and $\hat{y}_{i}$ are the ground truth and estimated number of people corresponding to the $i^{th}$ sample, which is obtained by summing all the pixel values in the density map.

\textbf{Density Map Quality.}\quad To further evaluate the quality of density maps, we also calculate PSNR (Peak Signal-to-Noise Ratio) and SSIM (Structural Similarity in Image) \cite{wang2004image}. For those two metrics, the larger the value, the higher the quality of the predicted density maps.

\subsection{Datasets}
The synthetic dataset GCC \cite{wang2019learning} is the only source domain in this paper. As for the target domain, to ensure the sufficiency of our experiments, we respectively select two datasets from high-level density, medium-level density and low-level density datasets, a total of six datasets, namely UCF-QNRF \cite{idrees2018composition}, Shanghai Tech Part A \cite{zhang2016single}, Shanghai Tech Part B \cite{zhang2016single}, WorldExpo'10 \cite{zhang2016data} , Mall \cite{chen2012feature} and UCSD \cite{chan2008privacy}.

\textbf{Source Domain Dataset.}\quad \emph{GCC} is a large-scale synthetic dataset, which is sampled from $400$ virtual scenes by a computer mod. It contains $15,212$ of accurately annotated images with a total of $7,625,843$ instances. There is an average of $501$ people in each image.

\textbf{Congested Crowd Dataset.}\quad \emph{UCF-QNRF} is collected from a shared image website. Therefore, the dataset contains a variety of scenes. It consists of $1,535$ images($1201$ training  and $334$ testing images), with $1,251,642$ annotated instances. The average number of people is 815 per image.
\emph{Shanghai Tech Part A} is also randomly collected from the Internet with different scenarios. It consists of $482$ images  ($300 $ training and $182$ testing images) with different resolutions. The average number of people in an image is $501$.

\textbf{Moderate Crowd Dataset.}\quad \emph{Shanghai Tech Part B} is captured from the surveillance camera on the Nanjing Road in Shanghai, China. It contains $716$ samples ($400$ training and $316$ testing images). The scenes are relatively consistent, with an average of $123$ people per picture. \emph{WorldExpo'10} consists of $3,980$ labeled images. They are collected from $108$ surveillance scenes ($103$ scenes for training and the remaining $5$ scenes for testing) in Shanghai $2010$ WorldExpo. The average number of people is 50  per image.

\textbf{Sparse Crowd Dataset.}\quad \emph{Mall} is captured from a surveillance camera installed in a shopping mall, which records the $2,000$ ($800$ for training and $1,200$ for testing) sequential frames.
The average of people in each image is $31$.
\emph{UCSD} consists of $2,000$ frames (frames $601-1,400$ for training and the others for testing), which are collected from a single-scene surveillance video. The average number of pedestrian in each image is $25$.

\subsection{Ablation Study}\label{sec:ablation}
We present our ablation experiments from two perspectives. Firstly, regarding the few-shot data, we demonstrate the impact by using different training methods. Secondly, for the proposed NLT, we discuss the effects of $\theta^{f}$ and $\theta^{b}$  on modeling the domain shift. The following experiments are conducted on Shanghai Tech Part A and B datasets, and the selected few-shot data are both the $10\%$ of the training set.
\begin{table*}[htbp]
	\centering
	\caption{The effectiveness of the domain shift parameters $\theta^{f}$ and $\theta^{b}$ on the testing set
of Shanghai Tech Part A and B.}
	\setlength{\tabcolsep}{9pt}{
		\begin{tabular}{c|ccIc|c|c|cIc|c|c|c}
			\whline
			\multirow{2}{*}{Method}	&\multirow{2}{*}{DA} &\multirow{2}{*}{FS} &\multicolumn{4}{cI}{Shanghai Tech Part A} &\multicolumn{4}{c}{Shanghai Tech Part B}\\
			\cline{4-11}
			&&& MAE &MSE &PSNR &SSIM &MAE & MSE &PSNR &SSIM \\
			\whline
			NoAdpt                      &\xmark &\xmark &188.0 &279.6 &20.91 &0.670  &20.1 &29.2 &26.62 &0.895\\
			\hline
			Fine-tuning                 &\rmark &\rmark &105.7 &167.6 &21.72 &0.702  &13.8 &22.3 &27.0 &0.931    \\
            \whline
			Factor ($\theta^{f}$)      &\rmark &\rmark &109.2 &161.3 &21.49 &0.758     &13.5 &23.5 &27.26 &0.921 \\
			\hline	
			bias ($\theta^{b}$)        &\rmark &\rmark &107.8 &169.9 &21.14 &0.796   &12.8 &20.6 &27.17 &0.916 \\
			\hline	
			NLT ($\theta^{f}+\theta^{b}$)  &\rmark  &\rmark &\bfseries93.8 &\bfseries157.2 &\bfseries21.89 &\bfseries0.729
                                            &\bfseries11.8 &\bfseries19.2  &\bfseries27.58 &\bfseries0.937  \\
			\whline
		\end{tabular}
	}
\label{Table:parameters}
\end{table*}

\textbf{Compared with Other Training Methods.}\quad Six training methods are used to demonstrate the role of few-shot data in narrowing the domain gap. The specific settings are as follows:
\begin{itemize}
  \item \textbf{NoAdpt.}\quad Train the model on the GCC dataset.
  \item \textbf{Supervised.}\quad Train the model on few-shot data.
  \item \textbf{Fine-tuning a.}\quad Train the model on the GCC dataset and fine-tune all parameters with few-shot data.
  \item \textbf{Fine-tuning b.}\quad Train the model on the GCC dataset and fine-tune the decoder (last four layers) with few-shot data.
  \item \textbf{NLT (ours).}\quad Train the model from GCC to the real-world dataset with our NLT and training strategy.
  \item \textbf{IFS+NLT (ours).}\quad Replace the original GCC data with IFS \cite{gao2019domain} translated GCC \cite{wang2019learning} in the last setting.
\end{itemize}

As shown in Fig. \ref{fig:methods}, we draw the loss and performance curves on the validation set during training. Taking Shanghai Tech Part A dataset as an example, it is difficult to reduce the loss of the validation set without domain adaptation. The supervised training and fine-tuning with few-shot data can significantly reduce the loss, but it is easy to suffer from overfitting. Compared to supervised training and fine-tuning, NLT can reach lower validation loss and inhibit overfitting.
In Fig. \ref{fig:methods} (b) and (c), the MAE and MSE curves also illustrate the effectiveness of NLT. Similarly, in Fig. \ref{fig:methods} (d), (e) and (f), Shanghai Part B have the same trend, which proves that NLT is effective for both dense and sparse scenes.

Table \ref{Table:methods} shows the testing results of six training
methods. No adaptation is usually hard to achieve satisfying results, which validates the vast distance between real and synthetic data. As shown in lines 3, 4, and 5, fully supervised training and fine-tuning on a pre-trained GCC model with few-shot data yield better results than no adaptation. It shows that few-shot data has a significant effect on narrowing the domain gap. By comparing the results of lines 4, 5, and 6, the proposed NLT yields better performance than the fine-tuning operation. Taking MAE as an example, in Shanghai Tech part A, NLT reduces the counting error by {\color{red}{13.1\%}} compared with fine-tuning a. In Shanghai Tech part B, it reduces by {\color{red}{10.0\%}}. We also test NLT on the stylized images generated by IFS \cite{gao2019domain}, and the results show that NLT can help other domain adaptation methods to improve performance further. In conclusion, the proposed NLT maximizes the potential of few-shot data.
\begin{table*}[htbp]
	\centering
    \small
	\caption{The performance of other domain adaptation (DA) methods and the proposed NLT on the six real-world datasets. FS refers to $10\%$ shot data from the target domain.}
	\setlength{\tabcolsep}{3.9pt}{
		\begin{tabular}{c|c|ccIc|c|c|cIc|c|c|cIc|c|c|c}
			\whline
			\multirow{2}{*}{Method}&\multirow{2}{*}{Backbone}&\multirow{2}{*}{DA}&\multirow{2}{*}{FS}
            &\multicolumn{4}{cI}{Shanghai Tech Part A}&\multicolumn{4}{cI}{Shanghai Tech Part B} &\multicolumn{4}{c}{UCF-QNRF}\\
		\cline{5-16}
		&&& & MAE &MSE &PSNR &SSIM &MAE & MSE &PSNR &SSIM &MAE & MSE &PSNR &SSIM \\
		\whline
		CycleGAN \cite{zhu2017unpaired}&VGG-16&\rmark &\xmark &143.3 &204.3 &19.27 &0.379    &25.4 &39.7
                                                &24.60 &0.763    &257.3 &400.6 &20.80 &0.480 \\
		\hline	
		SE CycleGAN \cite{wang2019learning}&VGG-16&\rmark &\xmark&123.4 &193.4 &18.61 &0.407  &19.9 &28.3
                                                    &24.78 &0.765    &230.4 &384.5 &21.03 &0.660 \\
		\hline	
        FA \cite{gao2019feature} &VGG-16 &\rmark &\xmark  &-  &-  &- &- &16.0 &24.7 &- &- &- &-  &-  &-\\
        \hline
        FSC \cite{han2020focus}&VGG-16 &\rmark &\xmark &129.3 &187.6 &21.58 &0.513 &16.9 &24.7 &26.20
                                                        &0.818   &221.2 &390.2 &\bfseries23.10&0.7084\\
        \hline
         IFS \cite{gao2019domain}&VGG-16&\rmark &\xmark &112.4 &176.9 &21.94 &0.502  &13.1 &19.4 &\bfseries28.03 &0.888   &211.7 &357.9 &21.94 &0.687\\
         \whline
		NoAdpt (ours) &VGG-16 &\xmark &\xmark   &188.0 &279.6 &20.91 &0.670 & 20.1 &29.2 &26.62 &0.895
                                                &276.8 &453.7 &22.22 &0.692  \\
		\hline
	\rowcolor{Lavender}	NLT (ours)&VGG-16&\rmark &\rmark
                    &93.8  &157.2 &21.89 &0.729
                    &11.8  &19.2 &27.58 &0.937
			        &172.3 &307.1 &22.8 &0.729\\

	\rowcolor{Lavender}	IFS\cite{gao2019domain}+NLT (ours) &VGG-16&\rmark &\rmark
                    &\bfseries90.1 &\bfseries151.6 &\bfseries22.01 &0.741
                    &10.8          &\bfseries18.3 &{27.69} &{0.932}
			        &\bfseries157.2 &\bfseries263.1 &{23.01} &\bfseries0.744\\
	\whline
    \rowcolor{LightCyan}	NLT (ours)&ResNet-50&\rmark &\rmark
                    &{91.4} &{153.4} &{21.45} &\bfseries{0.749}
                    &\bfseries{10.4}  &{18.8} &{27.79} &\bfseries0.942
			        &{165.8} &{279.7} &{22.89} &{0.734}\\
    \whline
		\end{tabular}
		\vspace{0.12cm}
	}
    \setlength{\tabcolsep}{3.3pt}{
	\begin{tabular}{c|c|ccIc|c|c|c|c|cIc|c|c|cIc|c|c|c}
		\whline
		\multirow{2}{*}{Method}&\multirow{2}{*}{Backbone}&\multirow{2}{*}{DA} &\multirow{2}{*}{FS} &\multicolumn{6}{cI}{WorldExpo'10 (only MAE)} &\multicolumn{4}{cI}{UCSD} &\multicolumn{4}{c}{MALL}\\
		\cline{5-18}
		&&&&S1 &S2 &S3 &S4 &S5 &Avg. & MAE &MSE &PSNR &SSIM & MAE &MSE &PSNR &SSIM \\
		\whline
		CycleGAN \cite{zhu2017unpaired}&VGG-16 &\rmark &\xmark  &4.4 &69.6 &49.9 &29.2  &9.0  &32.4   &-
                                        &-    &-     &-   &-   &-    &-     &-  \\
		\hline	
		SE CycleGAN \cite{wang2019learning}&VGG-16 &\rmark &\xmark  &4.3 &59.1 &43.7  &17.0 &7.6  &26.3
                                                    &-   &- &-     &-   &-   &-    &-     &- \\
        \hline
		FA \cite{gao2019feature}&VGG-16 &\rmark &\xmark  &5.7 &59.9 &19.7 &\bfseries14.5 &8.1 &21.6   &2.0
                                                         &2.43 &- &-    &2.47 &3.25 &- &- \\
        \hline
        IFS  \cite{gao2019domain}&VGG-16 &\rmark &\xmark &4.5 &33.6 &\bfseries14.1  &30.4 &4.4  &17.4
                                            &1.76 &2.09 &24.42 &0.950 &2.31 &2.96 &25.54 &0.933 \\
		\whline
		NoAdpt (ours)&VGG-16 &\xmark &\xmark  &5.0 &89.9 &63.1  &20.8 &17.1 &39.2   &12.79 &13.22 &23.94
                                                &0.899 &6.20 &6.96 &24.65 &0.879 \\
		\hline
\rowcolor{Lavender} NLT (ours)&VGG-16&\rmark &\rmark   &{2.3} &{22.8} &{16.7} &{19.7}&{3.9}&{13.1}
		                                               &{1.58}  &{1.97} &{25.29}  &{0.942}
                                                       &{1.96} &{2.55} &{26.92} &\bfseries0.967 \\

\rowcolor{Lavender} IFS\cite{gao2019domain}+NLT (ours) &VGG-16&\rmark &\rmark
                                &\bfseries2.0 &\bfseries15.3 &{14.7} &{18.8}&3.4&\bfseries10.8
		                        &1.48  &1.81 &\bfseries25.58  &\bfseries0.965
                                &1.86 &\bfseries2.39 &\bfseries27.03 &{0.944} \\
\whline
\rowcolor{LightCyan}  NLT (ours) &ResNet-50&\rmark &\rmark &3.1 &17.8 &{17.9} &{20.6}&\bfseries3.2&12.5
		                                                   &\bfseries1.42 &\bfseries1.76 &25.56 &0.964
                                                           &\bfseries1.80 &2.42 &26.84 &{0.940} \\
\whline
\end{tabular}
\label{Table-DA}
}
\end{table*}

\textbf{The Influence of Domain Shift Parameters.}\quad
For the proposed NLT, two types of parameters are defined to learn the transformation of neurons, namely $\theta^{f}$ and $\theta^{b}$. To verify the parameters' validity and compatibility, we use factor $\theta^{f}$, bias $\theta^{b}$, and both of them to model the model shift, respectively. The details of the experiments are shown in Fig .\ref{fig:parameters}.

The red curves represent no domain adaptation results. The loss curve is descending at the beginning of the training. However, as time goes on, it changes from decrease to increase. The reason is that the model trained with synthetic data has a limited ability to fit the real data. Once the limit value is passed, the model will continuously deviate from the target domain. The blue and green curves show the effectiveness of domain factor $\theta^{f}$ and domain bias $\theta^{b}$, respectively, both of them can significantly reduce losses and improve performance. It is worth noting that factor is not easy to overfit, but the convergence is slow, while bias converges faster but is easy to overfit. When the two are together, they complement each other and perform best.

The results of the test set are shown in Table \ref{Table:parameters}. The learnable parameters for factors $\theta^{f}$ and bias $\theta^{b}$ both are $\sim1/9$ of the source model. Fine-tuning updates all parameters of the source model. For example, in Shanghai Tech Part A, $10\%$ of the training set are treated as few-shot data, factor $\theta^{f}$ and bias $\theta^{b}$ achieve the similar results compared with fine-tuning, which verify that it is effective to use factor and bias to represent domain shift. We achieve the best results when combining them as NLT.

\subsection{Adaptation Results on Real-world Datasets}
In this section, we test the performance of the NLT by using it to learn the domain shift from GCC to six real-world datasets and compare it with the other domain adaptation methods.

\textbf{Metrics Report.}\quad Table \ref{Table-DA} lists the statistical results of the four metrics (MAE$\downarrow$/MSE$\downarrow$/PSNR$\uparrow$/SSIM$\uparrow$). Compared with the image translation (CycleGAN \cite{zhu2017unpaired}, SE CycleGAN \cite{wang2019learning}, and IFS \cite{gao2019domain}) and feature adversarial learning (FA \cite{gao2019feature} and FSC \cite{han2020focus}) methods, our method performs better with the use of $10\%$ annotated data in the target domain. Taking MAE as an example, as the lavender row shows, NLT reduce counting errors by {\color{red}{16.5\%}}, {\color{red}{10.0\%}}, {\color{red}{18.6\%}}, {\color{red}{29.9\%}}, {\color{red}{10.0\%}}, and {\color{red}{15.2\%}} compared with the above methods on six real-world datasets, respectively. As for the quality of predicted density map, we also achieve a significant improvement on PSNR and SSIM, indicating that few-shot data make a great contribution to noise cancellation in the background region. Experiments with different density datasets demonstrate the universality of NLT for cross-domain counting tasks.

Furthermore, we discuss the combination of NLT and other domain adaptation methods. We implement stylistic realism for the GCC \cite{wang2019learning} dataset by using IFS \cite{gao2019domain}, which is a image translation method for cross-domain crowd counting. These images are then treated as source domain data, and the proposed NLT is applied to achieve the domain adaptation. The final test results in the six real-world datasets are shown in Table \ref{Table-DA}, lavender row. Compared with the original IFS \cite{gao2019domain}, the NLT decreases the MAE by {\color{red}{19.8\%}}, {\color{red}{17.6\%}}, {\color{red}{25.7\%}}, {\color{red}{37.9\%}}, {\color{red}{16.0\%}}, and {\color{red}{19.5\%}} on the six real data sets, respectively. We also test NLT on the ResNet50 architecture, and the test results are shown in Table \ref{Table-DA}, light cyan row. On the whole, the ResNet50 backbone outperforms the VGG-16 backbone because of its deeper structure and superior semantic extraction capability.
\begin{figure*}[htbp]
\centering
\includegraphics[width=0.95\textwidth]{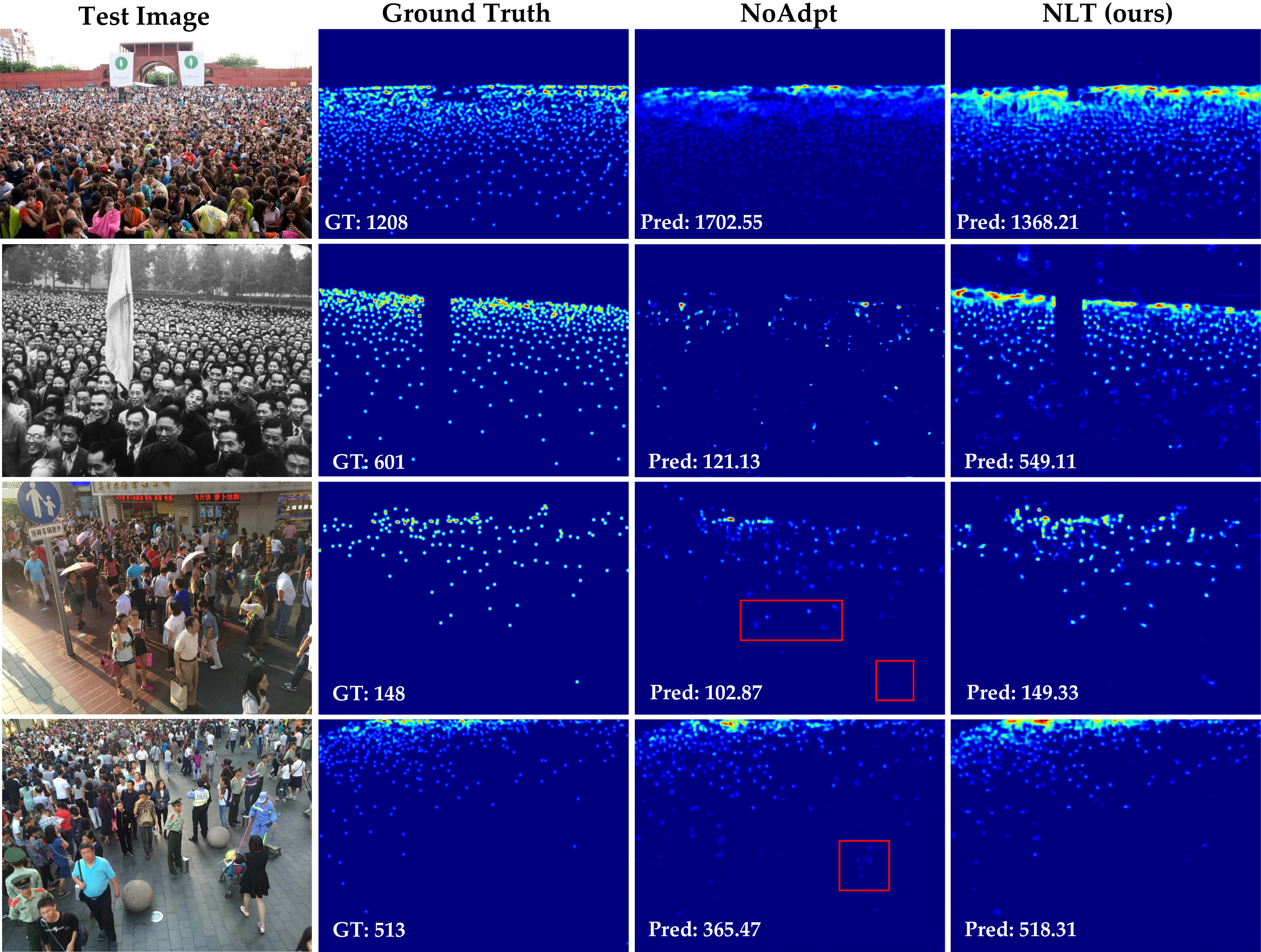}
\caption{Exemplar results of adaptation from GCC to Shanghai Tech Part A/B
dataset. Row 1 and 2 come from Shanghai Tech Part A, and others
are from Part B.}
\label{fig:result}
\end{figure*}

\begin{figure*}[htbp]
\centering
\includegraphics[width=0.98\textwidth]{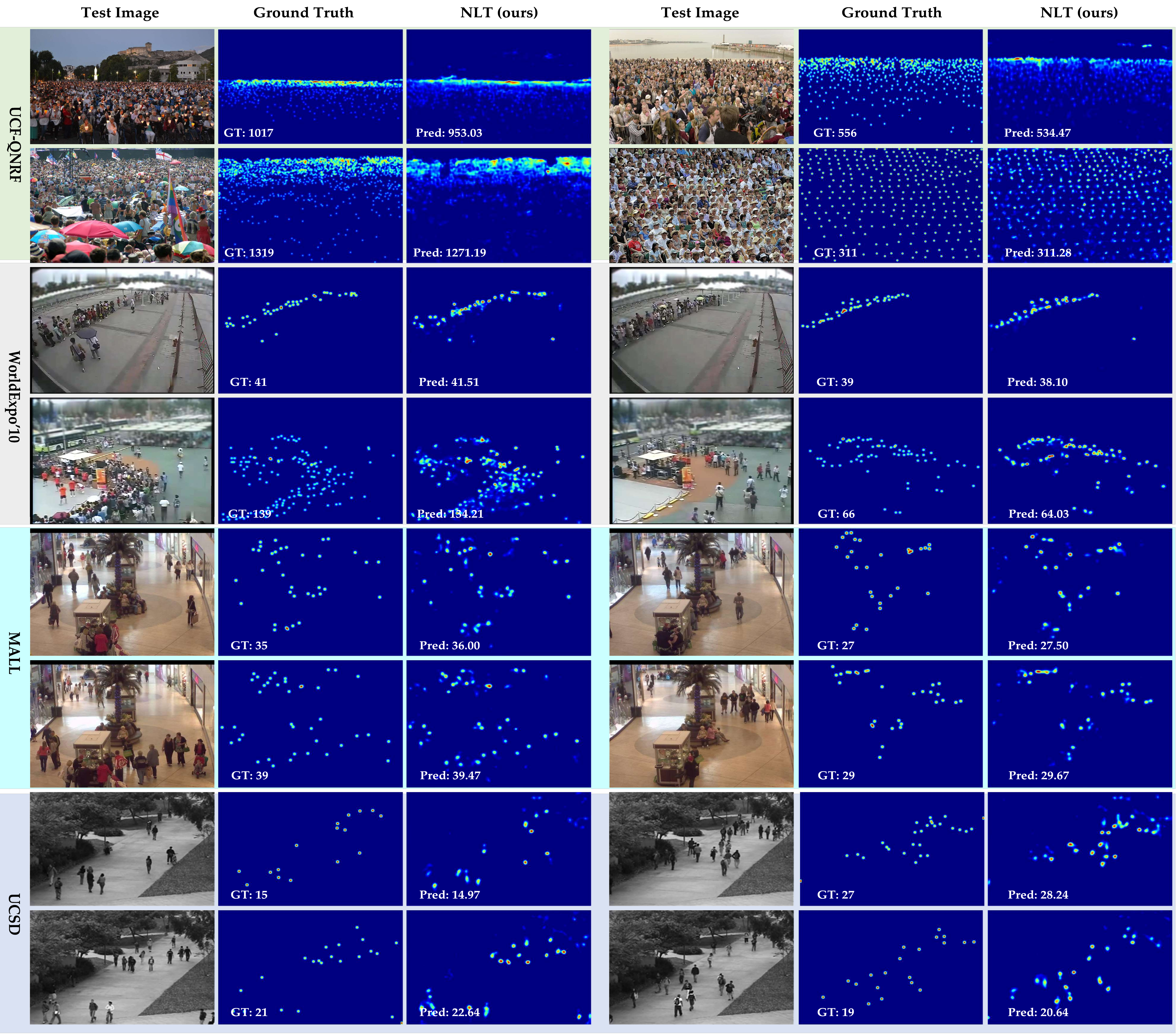}
\caption{More visual samples of adaptation from GCC to other four real-world datasets with our proposed NLT.}
\label{fig:more_result}
\end{figure*}

\textbf{Visualization Results.}\quad Fig. \ref{fig:result} shows the visualization results of no adaptation and the proposed NLT. Column 3 shows the results without domain adaptation. The regression results are not acceptable in a congested scene like Shanghai Tech Part A, especially the gray-scale image in Row 2. On Shanghai Tech Part B, the counting results of no domain adaptation are a little close to the ground truth, but the problems remain in detail and background. The red box in Row 3 shows that the regression value is weak in some areas. The estimation errors in the background region also prevent the performance, such as the red box shown on Raw 4. After the domain adaptation, the above questions are alleviated. In general, the NLT improves the density map in counting values and details.
To more intuitively demonstrate our domain adaptation effect, we show more results in Fig. \ref{fig:more_result}. From the performance on different datasets, NLT is effective for cross-domain counting tasks with different crowding levels.

\textbf{Computational Statistics.}\quad Table \ref{Table:GFLOPs} shows the computation statistics on the proposed NLT and several other DA methods, including parameter size, Giga Floating-point Operations Per second (GFLOPs), and Frames Per Second (FPS). All the methods are tested by inputting images with a size of 768x1024 under a single GTX-1080Ti GPU. Since SE CycleGAN \cite{wang2019learning} and IFS \cite{gao2019domain} involve image transformation in the training phase, we only compute the crowd counters in all methods here. All crowd counters use VGG-16 as the backbone. NLT ranks below the IFS because the NLT introduces the domain shift parameters $\theta^{f}$ and $\theta^{b}$. The advantage of NLT is that the training process is faster as IFS needs image transfer.
\begin{table}[htbp]
	\centering
	\small
	\caption{The computational statistics of the proposed NLT and other domain adaptation methods.}
	\setlength{\tabcolsep}{3.5pt}{
		\begin{tabular}{c|c|c|c|c}
			\whline
			\multirow{2}{*}{Method}	&SE CycleGAN\cite{wang2019learning} &\multirow{2}{*}{FSC\cite{han2020focus}} &\multirow{2}{*}{IFS\cite{gao2019domain}} & \multirow{2}{*}{NLT(ours)}\\

            &FA\cite{gao2019feature} &&&\\
            \hline
			params(MB)      &67.1  &68.6  &\bfseries30.1  & 38.1\\
			\hline
			GFLOPs          &326.8 &311.8 &\bfseries236.9 & 278.9\\
			\hline
			FPS             &9.3   &9.6   &\bfseries18.3  & 16.7\\
            \whline
		\end{tabular}	
	}
\label{Table:GFLOPs}
\end{table}

\subsection{Statistical Analysis of Domain Shift}
\begin{figure*}[htbp]
\centering
\includegraphics[width=0.95\textwidth]{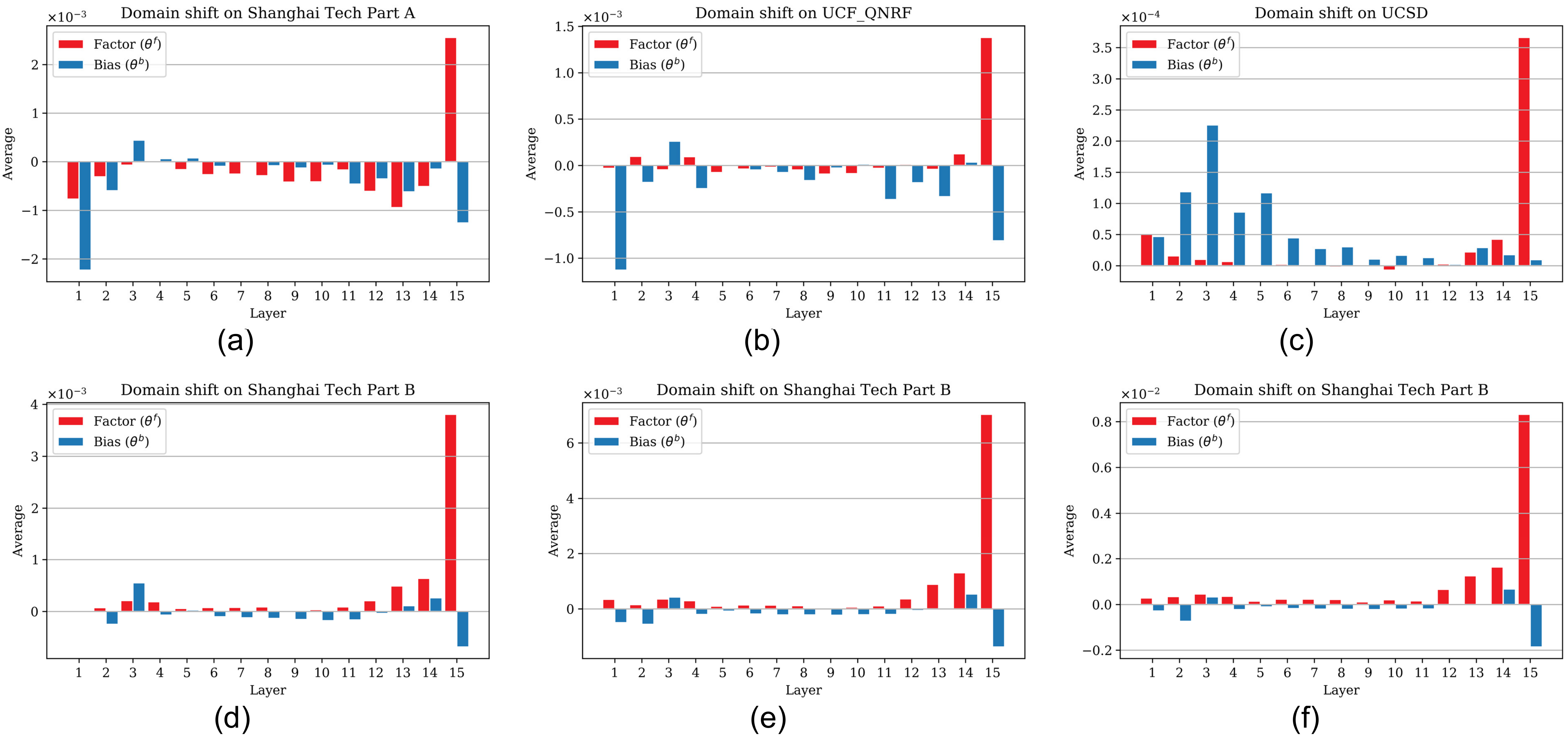}

\caption{The averages of domain factor and domain bias in each layer of the network.}
\label{fig:domainshift}
\end{figure*}
Domain factor  $\theta^{f}$ and domain bias  $\theta^{b}$ are defined as the parameters to model the shift from the source domain to the target domain. They are initialized to 1 and 0, respectively, and optimized to narrow the domain gap by few-shot data. In Sec. \ref{sec:ablation}, we verify its effectiveness by testing it on different datasets. In this section, we will further analyze the significance of these parameters from mathematical statistics.

There are $15$ convolutional layers in the VGG-16 backbone and the decoder, and each convolution kernel contains a domain factor and domain bias parameter. For the well-trained target model, we calculate the mean values of factor and bias at each layer. The statistical results are shown in Fig. \ref{fig:domainshift}, where the mean value for factor is subtracted from the initial value $1$. As Fig. \ref{fig:domainshift} (a) shown, at most layers, the mean value of factor and bias are less than $1$ and $0$, respectively. Therefore, the effect of factor $\theta^{f}$ and bias  $\theta^{b}$ is to shrink the parameters of the source model. We call this shift a ``down domain shift". The distribution of UCF-QNRF in Fig. \ref{fig:domainshift} (b) is similar to that of Shanghai Tech Part A. Both of them are collected from the internet, so it has a similar distribution. In Fig. \ref{fig:domainshift} (c), the averages of factor and bias are greater than $1$ and $0$ in most layers, respectively. We define
the shift as ``up domain shift." In Fig. \ref{fig:domainshift} (d), factor and bias are distributed on both sides. We define this shift as ``up-down domain shift". In addition, in Fig. \ref{fig:domainshift} (d)(e)(f), we use $10\%$, $30\%$, and $50\%$ of the training set as few-shot data to learn domain shift parameters. The distributions are the same eventually. This reveals that only a few target domain labeled images are needed to learn the representation of domain shift.

\subsection{Analysis in Selecting Few-shot Data}
\begin{figure}[!htbp]
\centering
\includegraphics[width=0.5\textwidth]{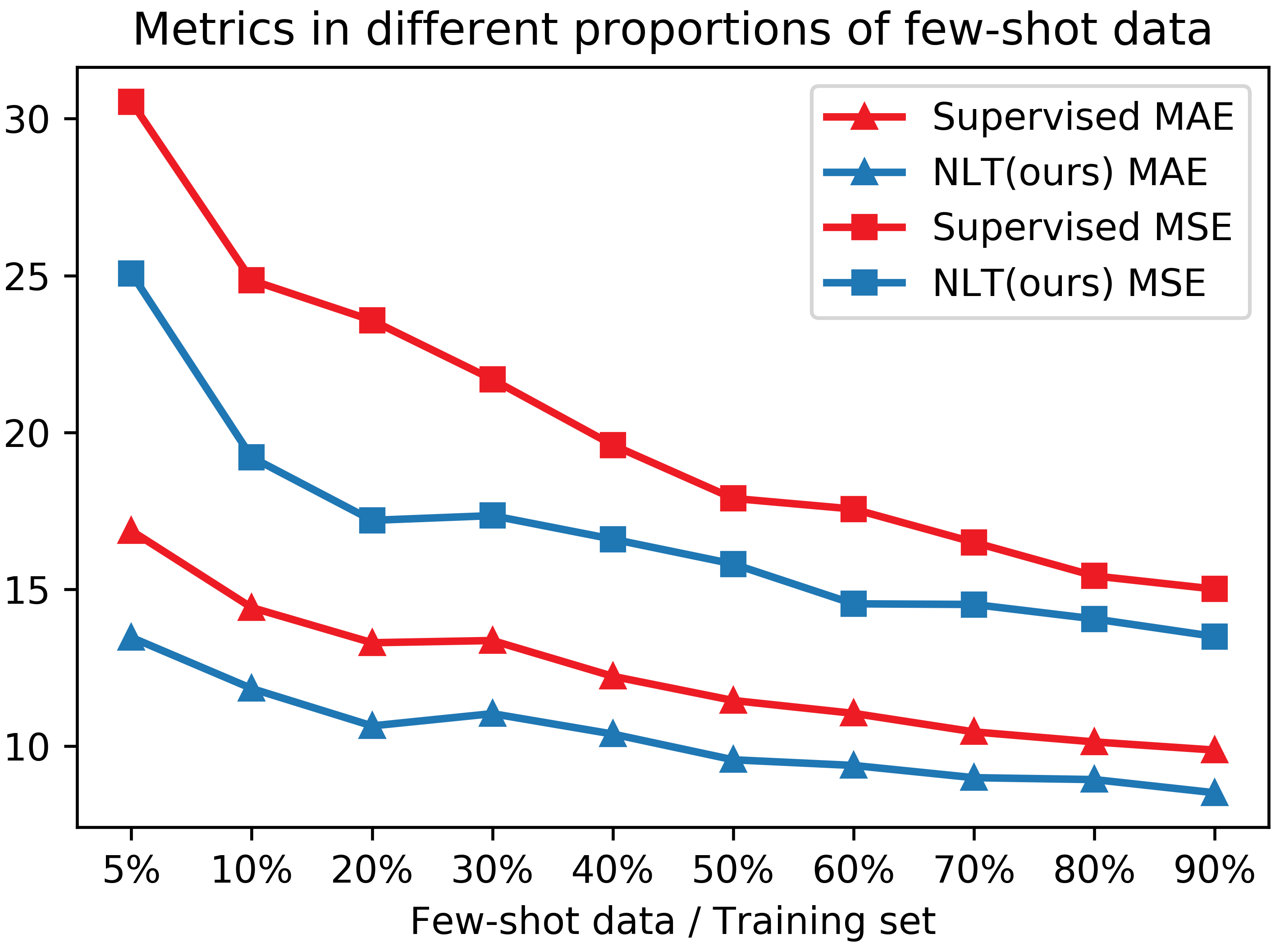}
\caption{The testing results for NLT (blue) and supervised training (red) with different ratios of few-shot data on Shanghai Tech Part B. The triangle and rectangle represent MAE and MSE, respectively.}
\label{Fig:select}
\end{figure}

Since our domain adaptation method requires a few target domain labeled images, in this section, we will discuss the effects of selecting different proportions of few-shot data for NLT. As shown in Fig. \ref{Fig:select}, we conduct the experiments on the Shanghai  Tech Part B dataset. The horizontal axis represents different proportions of the few-shot images. The vertical axis represents the metric values on the Shanghai Tech Part B testing set. The experiments illustrate that NLT performs better with increasing few-shot learning data.  Besides, compared with the traditional supervised training methods, the proposed NLT is better in every configuration. It shows that NLT can improve the generalization ability of the supervised model. Overall, the NLT is robust in the different shot data settings.

\section{Conclusions}
This paper summarizes the existing problems of CDCC methods and rethink the domain shift from model-level. To convert the source model to the target model, we propose a Neuron Linear Transformation (NLT) method to model the domain shift and optimize the domain shift parameters by few-shot learning. Extensive experiments show that the NLT achieves comparable performance with other domain adaptation methods by using $10\%$ target domain data. Besides, it also has a better expression ability for domain shift. The proposed NLT also be applied in other domain adaptation tasks in future work, such as semantic segmentation, pedestrian Re-ID, and saliency object detection. Take the saliency object detection task \cite{wang2019inferring,wang2018salient,wang2017deep} as an example, it is similar to binary segmentation task, which needs either bounding boxes or pixel-level annotations to supervise. We can also utilize computer mod to synthesize numerous images and annotate the saliency region automatically. The NLT proposed in this paper can be employed to model the domain gap of cross-domain saliency object detection.


\bibliographystyle{IEEEtran}
\bibliography{IEEEabrv,reference}

\end{document}